# Eigenlogic: Interpretable Quantum Observables with applications to Fuzzy Behavior of Vehicular Robots

*Zeno Toffano [i], François Dubois [ii]*

## 1. Outline

This work proposes a formulation of propositional logic, named Eigenlogic, using quantum observables as propositions. The eigenvalues of these operators are the truth-values and the associated eigenvectors the interpretations of the propositional system. Fuzzy logic arises naturally when considering vectors outside the eigensystem, the fuzzy membership function is obtained by the Born rule of the logical observable.
This approach is then applied in the context of quantum robots using simple behavioral agents represented by Braitenberg vehicles. Processing with non-classical logic such as multivalued logic, fuzzy logic and the quantum Eigenlogic permits to enlarge the behavior possibilities and the associated decisions of these simple agents.

## 2. History: Boole: "0" and "1"; von Neumann: quantum projections as propositions

George Boole gave a mathematical symbolism through the two numbers {0,1} representing *resp.* the "false" or "true" character of a proposition (Boole 1847). An *idempotent* symbol $x$ verifies the equation: $x^2 = x$, which admits only two possible values : 0 and 1. This equation was considered by Boole as the "fundamental law of thought", the associated formulation for logic is operational as pointed out in (Halperin 1981) because $x$ acts as a selection operator on classes. As will be emphasized here the algebra of idempotent symbols can also be interpreted as an algebra of commuting projection operators and used for developing propositional logic in a quantum linear-algebraic framework (Toffano 2015).

John von Neumann (Von Neumann 1932) considered projectors as propositions, he also introduced the formalism of the density matrix in quantum mechanics where a pure quantum state $|\psi\rangle$ is be represented by a rank-1 projection operator: $\hat{\rho} = |\psi\rangle\langle\psi|$.

[i] CentraleSupélec, Telecom Dep., Gif-sur-Yvette, Laboratoire des Signaux et Systèmes, L2S - UMR8506-CNRS, Université Paris-Saclay, France.
[ii] AFSCET (French Association for Systems Science) and , Department of Mathematics, Conservatoire National des Arts et Métiers, Paris, France.          [WOSC, Rome, January 25-27 2017,     subm: March 29, 2017]

## 3. Eigenlogic: quantum observable logic

A projection operator in Hilbert space is associated to a logical proposition, the operator being Hermitian it has the properties of a *quantum observable* and is considered here a logical observable. This view is named *Eigenlogic* (Toffano 2015), (Dubois & Toffano 2017) and can be summarized:

eigenvectors in Hilbert space ⟺ interpretations (atomic propositional cases)
logical observables ⟺ logical connectives
eigenvalues ⟺ truth values

One can express the logical observable as a development:
$$\mathbf{F} = f(0)\mathbf{\Pi}_0 + f(1)\mathbf{\Pi}_1 = diag(f(0), f(1))$$
the terms of the development are the 2-dimensional rank-1 projectors $\mathbf{\Pi}_0$ and $\mathbf{\Pi}_1$, the cofactors $f(0)$ and $f(1)$ are the eigenvalues and correspond to the truth values $\{0,1\}$ of the logical connective. This allows to generate 4 one-argument logical observables:
$$\mathbf{F}_A = 0\mathbf{\Pi}_0 + 1\mathbf{\Pi}_1 = \mathbf{\Pi} \;,\; \mathbf{F}_{\bar{A}} = 1\mathbf{\Pi}_0 + 0\mathbf{\Pi}_1 = \mathbf{I}_2 - \mathbf{\Pi}\;,\; \mathbf{F}_\emptyset = 0\mathbf{\Pi}_0 + 0\mathbf{\Pi}_1 = \mathbb{O}_2\;,\; \mathbf{F}_U = 1\mathbf{\Pi}_0 + 1\mathbf{\Pi}_1 = \mathbf{I}_2$$
Then the two-argument logical observables can be developed on the corresponding four rank-1 projectors:
$$\mathbf{F}_2 = diag[f(0,0), f(0,1), f(1,0), f(1,1)]$$
There are $2^{2^n}$ logical connectives for a $n$-argument (*arity*) system. For $n = 2$ this gives the 16 binary logical connectives *e.g.*: AND, OR ,XOR, →, ↔,...

The complete orthonormal basis for a two input quantum states are: $|00\rangle$, $|01\rangle$, $|10\rangle$ and $|11\rangle$. These state vectors are the eigenvectors of the logical obseervables and correspond to *interpretations* of the logical system. What happens when the quantum state is not one of the eigenvectors of the logical system?

In quantum mechanics one can always express a state vector as a combination on a complete orthonormal basis. In particular on the canonical eigenbasis of the logical observable family:
$$|\psi\rangle = c_{00}|00\rangle + c_{01}|01\rangle + c_{10}|10\rangle + c_{11}|11\rangle$$
When only one of the coefficients is not zero, then one has the case of a determined interpretation for the proposition.

*Fuzzy logic* (Zadeh 1965) deals with truth values that can take values between 0 and 1, so the truth of a proposition can lie between "completely true" and "completely false". When more than one coefficient in the development of $|\psi\rangle$ is non-zero one can give a "fuzzy" interpretation, and the quantum state $|\psi\rangle$ can be considered as a *quantum superposition of interpretations*.

For a projective observable $\mathbf{F}$ measured in the context of a quantum state, $|\psi\rangle$ the mean value (Born rule) gives directly a probability measure by:
$$p_{|\psi\rangle} = \langle\psi|\mathbf{F}|\psi\rangle = Tr(\boldsymbol{\rho} \cdot \mathbf{F}) \quad \text{with} \quad \boldsymbol{\rho} = |\psi\rangle\langle\psi| \quad \text{the } density\ matrix$$
The mean value of the logical projector observable $\mathbf{F}$ is thus a fuzzy measure of the truth of a logical proposition in the form of a fuzzy membership function $\mu$.

For one-argument an arbitrary 2-dimensional quantum state is:
$$|\phi\rangle = \sin\alpha|0\rangle + e^{i\beta}\cos\alpha|1\rangle$$
where $\alpha = \theta/2$ and $\beta = \varphi/2$ are real numbers and these angles can be represented on the *Bloch sphere* (see Fig. 1). The quantum mean value of the logical projector observable $\mathbf{A} = \mathbf{\Pi}$ is then given by:

$$\mu(a) = \langle\phi|\mathbf{\Pi}|\phi\rangle = \cos^2\alpha \qquad \text{representing a probability.}$$

*Figure 1: Bloch sphere with the general qubit quantum state |ϕ⟩ characterized by angles θ and φ.*

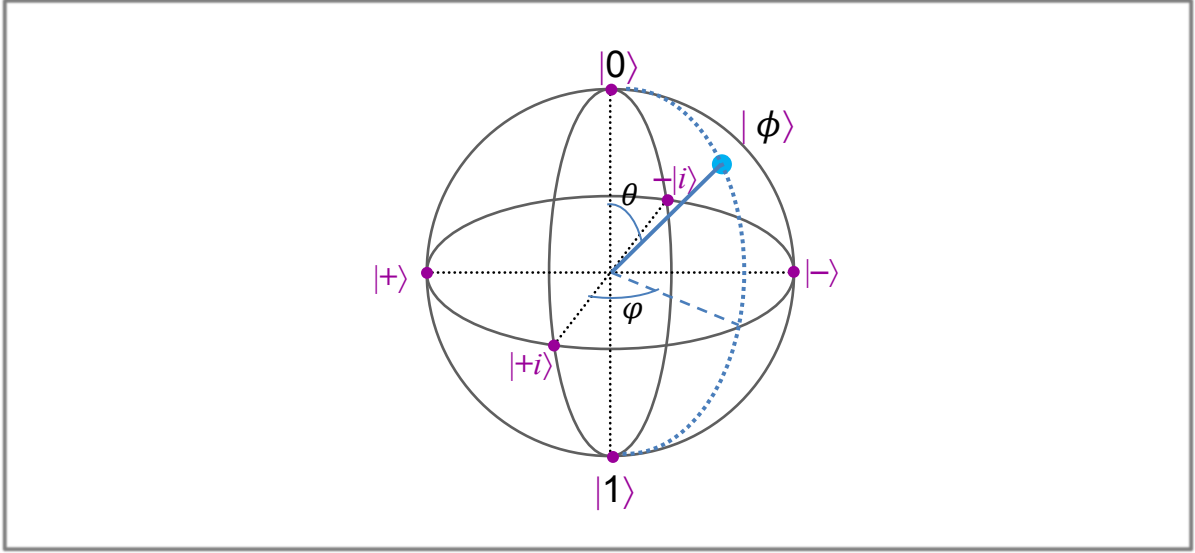

The recently observed revival of interest in applying *multi-valued logic* to the description of quantum phenomena is closely connected with fuzzy logic. Multi-valued logic is of interest to engineers involved in various aspects of information technology and has a long history in *computer aided design*.

The total number of logical connectives for a system of $m$ values and $n$ arguments is the combinatorial number: $m^{m^n}$. For a 3-valued 2-argument system: $3^{3^2} = 19683$. Showing that the possibilities of new connectives becomes intractable, but some special ones play an important role.

Multi-valued logic is naturally associated to quantum angular momentum: the eigenvalues of the $z$ component of the angular momentum observable for $l = 1$ is given by:

$$\boldsymbol{L_z} = \hbar \boldsymbol{\Lambda} = \hbar \, diag(+1, 0, -1)$$

with the associated logical truth values:

$$\text{"false"} \; F \equiv +1 \quad , \quad \text{"neutral"} \; N \equiv 0 \quad , \quad \text{"true"} \; T \equiv -1$$

The corresponding logical observables can be expressed as spectral decompositions over the three rank-1 projectors given by:
$$\boldsymbol{\Pi_{+1}} = \tfrac{1}{2}\boldsymbol{\Lambda}(\boldsymbol{\Lambda} + \mathbf{1}) \qquad \boldsymbol{\Pi_0} = \boldsymbol{I} - \boldsymbol{\Lambda^2} \qquad \boldsymbol{\Pi_{-1}} = \tfrac{1}{2}\boldsymbol{\Lambda}(\boldsymbol{\Lambda} - \mathbf{1})$$

**4. Eigenlogic applied to Quantum Robot Braitenberg Vehicles**

Valentino Braitenberg was a Cyberneticist and former director at the Max Planck Institute for Biological Cybernetics in Tübingen. A Braitenberg vehicle ($BV$) (Braitenberg 1986) is an agent that can autonomously move around based on its light sensor inputs. Depending on the sensor-motor wiring, it appears to achieve certain situations and to avoid others, changing course when situation changes. Several elementary vehicles can be considered:
- BV-2a (**fear**): turns away from the light if one sensor is activated more than the other.
- BV-2b (**aggress**): when the light source is placed near either sensor, the vehicle will go toward it.
- BV-3a (**love**): will go until it finds a light source, then slows to a stop.
- BV-3b (**explore**): goes to the nearby light source, but keeps an eye open to sail to a stronger source.

The vehicles can be designed according the *law of uphill analysis and downhill invention* (Braitenberg 1986), according to which it is far easier to create machines from simple structures that exhibit complex behavior than it is to try to build their structures from behavioral observations.

Practical realization of *BV*'s uses generally simple Boolean logic. It is interesting to extend the design to multi-valued, fuzzy or probabilistic logic and even quantum logic.

Paul Benioff (Benioff 1998) introduced the theoretical principle of a *quantum robot* as a first step towards a quantum mechanical description of systems that are aware of their environment and make decisions.

The research team of Marek Perkowski has designed robots, based on *BV*'s, using quantum gates and also introducing control, fuzziness and higher than binary valued logic (Raghuvanshi and Perkowski 2010). The potential applications presented here are inspired from these researches.

Considering the binary alphabet $\{+1,-1\}$ leads to analogies with the vehicle's behaviour, it has a natural correspondence with inhibition (negative: $-$) and excitation (positive : $+$). In this way the *BV*'s sensors *SL* and *SR* ( see Fig. 2) represent the inputs and the actuators *ML* and *MR* are represented by the 2-argument dictators $\boldsymbol{Z} = \text{diag}(1,1,-1,-1)$ and $\boldsymbol{Y} = \text{diag}(1,-1,1,-1)$.

The different possible combinations are given on Table 2.

*Table 1: quantum logical observables for BV actuators*

| Braitenberg Vehicle\ Actuator | ML | MR |
|---|---|---|
| BV2a (*fear*) | $-\boldsymbol{Z}$ | $-\boldsymbol{Y}$ |
| BV2b (*aggress*) | $-\boldsymbol{Y}$ | $-\boldsymbol{Z}$ |
| BV3a (*love*) | $+\boldsymbol{Z}$ | $+\boldsymbol{Y}$ |
| BV3b (*explore*) | $+\boldsymbol{Y}$ | $+\boldsymbol{Z}$ |

In another configuration the sensors *SL* and *SR* can be represented by tri-valued 2-argument dictators $\boldsymbol{U}$ and $\boldsymbol{V}$. For this purpose it is interesting to use the three positive values $\{0,1,2\}$ with he following interpretation:

$$\text{"no light"} \equiv 0 \qquad \text{"weak-level light"} \equiv 1 \qquad \text{"high-level light"} \equiv 2$$

Involved behaviors can thus be described using the *Min* and *Max* connectives. From the formulation given above based on the classical interpolation methods (Dubois and Toffano 2016) it is easy to derive, the expressions for the alphabet $\{0,1,2\}$, giving the following logical observables:

$$\boldsymbol{Min}_{3\,\{0,1,2\}} = \boldsymbol{U} + \boldsymbol{V} + \boldsymbol{U}^2\boldsymbol{V} + \boldsymbol{V}^2\boldsymbol{U} - \frac{1}{2}\boldsymbol{U}^2\boldsymbol{V}^2 - \frac{5}{2}\boldsymbol{UV} = diag(0,0,0,0,1,1,0,1,2)$$

$$\boldsymbol{Max}_{3\,\{0,1,2\}} = \frac{5}{2}\boldsymbol{UV} + \frac{1}{2}\boldsymbol{U}^2\boldsymbol{V}^2 - \boldsymbol{U}^2\boldsymbol{V} - \boldsymbol{V}^2\boldsymbol{U} = diag(0,1,2,1,1,2,2,2,2)$$

One could also combine the multivalued operators above in a fuzzy logic configuration, *i.e.* when using input states that are not eigenstates. Using fuzzy Eignelogic one can calculate the fuzzy membership functions for complement: $\mu(\bar{a})$, conjunction $\mu(a \wedge b)$ and disjunction $\mu(a \vee b)$. These functions could be implemented in the processor preceding the actuators *ML* and *MR* (see Fuzzy-Demux in Fig. 2).

*Figure 2: principle of a multivalued fuzzy quantum Braitenberg vehicle*

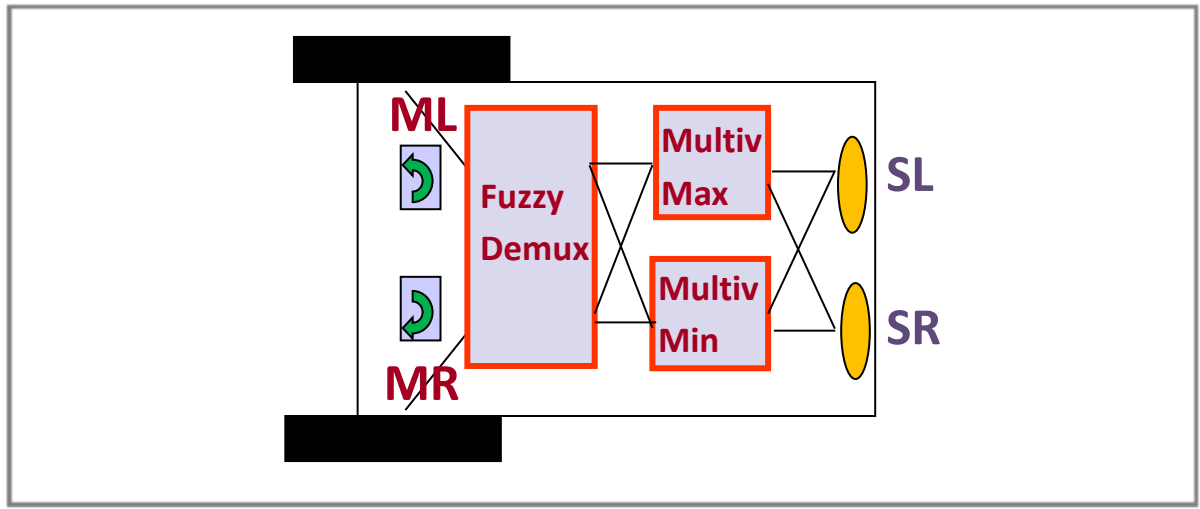

Fuzzy membership for logical implication (→) corresponding to motion-decision such as "step backwards", "step forwards", "turn left" and "turn right" can also be evaluated in this model. An idea presented in (Raghuvanshi and Perkowski 2010) consists in mapping emotions onto the Bloch sphere (Fig. 1) , this quantum fuzzy model uses conjunction, disjunction and complement operations. The internal emotional state could be described now by a quantum circuit built from counterparts of fuzzy operators. For instance, all kind of quantum phase gates could be used combined associated with recent techniques of *quantum tomography*.

**References**
Benioff, P., (1998)*,* "Quantum Robots and Environments"*, Physical Review A*, Vol. 58, No.2, 893–904.
Boole, G., (1847), *The Mathematical Analysis of Logic. Being an Essay To a Calculus of Deductive Reasoning*, Ed. Forgotten Books.
Braitenberg, V. (1986), Vehicles: Experiments in Synthetic, Psychology. MIT Press; Cambridge USA.
Dubois, F. and Toffano, Z., (2016): "Eigenlogic: a Quantum View for Multiple-Valued and Fuzzy Systems*",* Quantum Interaction 2016. *Lecture Notes in Computer Science*, vol 10106. Springer, 239-251, 2017, *arXiv*:1607.03509.
Halperin, T., (1981), "Boole's Algebra isn't Boolean Algebra. A Description Using Modern Algebra, of What Boole Really Did Create", *Mathematics Magazine*, Vol. 54, N0.4, 172–184.
Raghuvanshi, A. and Perkowski, M., (2010), "Fuzzy quantum circuits to model emotional behaviors of humanoid robots", Evolutionary Computation (CEC), IEEE Congr. on Evolutionary Computation, 18-23 July 2010, 1-8.
Toffano, Z., (2015): "Eigenlogic in the spirit of George Boole". *ArXiv*:1512.06632.
Von Neumann, J., (1932), *Mathematical Foundations of Quantum Mechanics*. Investigations in Physics, vol. 2, Princeton University Press, Princeton, (transl. 1955).
Zadeh, L.A., (1965), "Fuzzy sets"*, Information and Control*, 8 (3), 338-353.